\pdfoutput=1

\documentclass[11pt]{article}

\usepackage{EMNLP2023}

\usepackage{times}
\usepackage{latexsym}

\usepackage[T1]{fontenc}

\usepackage[utf8]{inputenc}

\usepackage{microtype}

\usepackage{inconsolata}

\usepackage{multirow}
\usepackage{graphicx}
\usepackage{arydshln}
\usepackage{booktabs}
\usepackage{graphicx}
\usepackage{color}
\usepackage{makecell}
\usepackage{url}
\title{LMStyle Benchmark: Evaluating Text Style Transfer for Chatbots}

\author{Jianlin Chen \\
  Carnegie Mellon University, Pittsburgh, PA \\
  \texttt{jianlinc@andrew.cmu.edu} \\}

\begin{document}
\maketitle
\begin{abstract}
Since the breakthrough of ChatGPT, large language models (LLMs) have garnered significant attention in the research community. With the development of LLMs, the question of text style transfer for conversational models has emerged as a natural extension, where chatbots may possess their own styles or even characters. However, standard evaluation metrics have not yet been established for this new settings. This paper aims to address this issue by proposing the LMStyle Benchmark, a novel evaluation framework applicable to chat-style text style transfer (C-TST), that can measure the quality of style transfer for LLMs in an automated and scalable manner. In addition to conventional style strength metrics, LMStyle Benchmark further considers a novel aspect of metrics called appropriateness, a high-level metrics take account of coherence, fluency and other implicit factors without the aid of reference samples. Our experiments demonstrate that the new evaluation methods introduced by LMStyle Benchmark have a higher correlation with human judgments in terms of appropriateness. Based on LMStyle Benchmark, we present a comprehensive list of evaluation results for popular LLMs, including LLaMA, Alpaca, and Vicuna, reflecting their stylistic properties, such as formality and sentiment strength, along with their appropriateness. 

\end{abstract}

\section{Introduction}

The recent success of ChatGPT~\citep{chatgpt} has garnered significant attention from both academia and industry. Since its emergence, the development of Large Language Models (LLMs) has skyrocketed, resulting in a plethora of powerful models~\citep{touvron2023llama,alpaca2023,vicuna2023,redpajama2023}. Concurrently, the demand for stylized chatbots has surged in tandem with the trend, driving research on Text Style Transfer (TST) for conversational LLMs~\citep{reif-etal-2022-recipe, suzgun-etal-2022-prompt, luo2023promptbased, roy2023conversation}. However, these works mainly focus on introducing new methods and do not comprehensively evaluate the style transfer capabilities of advanced LLMs, such as LLaMA \citep{touvron2023llama}, Alpaca \citep{alpaca2023}, and Vicuna \citep{vicuna2023}. Therefore, in this paper, we conduct a thorough evaluation of eleven different advanced LLMs on TST tasks. By examining the models' ability to transfer text styles, we can gain valuable insights into the advancements of language models.

Given our goal, a comprehensive automatic evaluation suite is of utmost importance. One potential alternative is to use evaluation methods from conventional text style transfer tasks, which typically center around text style transfer without any contexts. These traditional tasks bear a resemblance to translation tasks \citep{shen2017, prabhumoye-etal-2018-style, Fu_Tan_Peng_Zhao_Yan_2018, luo2019a,li-etal-2018-delete, wu2019, malmi-etal-2020-unsupervised, reid-zhong-2021-lewis} and can be referred to as translation-style text style transfer (T-TST) for simplicity. However, evaluating on these tasks is not our objective, as they are far away from many downstream applications.

On the contrary, current Large Language Models (LLMs) can generate style-transferred responses given contexts. We refer to this approach as chat-style text style transfer (C-TST), which can be easily adapted to many downstream applications. For example, this setting can be used to simulate conversations with various fictional characters \citep{han-etal-2022-meet} more effectively. Therefore, we place greater emphasis on this task setting.
To the best of our knowledge, only \citet{roy2023conversation} have previously worked on C-TST, but they employed the same evaluation metrics as T-TST. Some of these metrics, such as BLEU, have been shown to be inadequate for evaluating LLMs \citep{reif-etal-2022-recipe}. The core distinction between T-TST and C-TST renders traditional evaluation methods in text style transfer inappropriate for C-TST. As a consequence, there is currently no comprehensive automatic evaluation suite for C-TST.

To fill this gap, we introduce LMStyle Benchmark, a robust evaluation framework designed to evaluate LLM performance in C-TST tasks. By incorporating measurements of style strength and appropriateness, LMStyle Benchmark accounts for style, coherence, fluency, and other implicit factors given contexts. Moreover, the entire evaluation framework is fully automated, distinguishing it from human evaluation metrics such as those used by \citet{roy2023conversation}.

We summarize our main contributions as follows:
\begin{itemize}
    \item We propose LMStyle Benchmark, the first automatic evaluation framework for Chat-fashion text style transfer tasks.
    \item We introduce a novel metric called appropriateness, along with accurate methods to quantify its value. These methods involve ChatGPT and Negative Log Likelihood, which exhibit much higher correlation with human evaluation than traditional TST metrics.
    \item We present a comprehensive list of evaluation results for existing LLMs on C-TST, offering valuable insights into their stylistic attributes.
\end{itemize}

\section{Related Works}

Previous works have primarily evaluated TST tasks based on the style strength and semantic preservation, with some also including the Fluency dimension \citep{prabhumoye-etal-2018-style, suzgun-etal-2022-prompt, reif-etal-2022-recipe}.

In terms of evaluating \textbf{Style Strength}, the standard classifier-based method is dominant. \citet{dai-etal-2019-style} and \citet{goyal-etal-2021-multi} train classifiers with style labels based on fastText \citep{fasttext}. Classifiers based on TextCNN \citep{kim-2014-textcnn} are also widely employed \citep{shen2017, xu2018unpaired}. \citet{reid-zhong-2021-lewis} and \citet{roy2023conversation} train RoBERTa-base \citep{liu2019roberta} to evaluate style strength, while \citet{luo2023promptbased} employ RoBERTa-Large and SiEBERT \citep{HARTMANN202375}.

In terms of evaluating \textbf{Semantic Preservation}, previous approaches can be divided into two categories. The first category involves calculating BLEU score \citep{papineni-etal-2002-bleu} or revised BLEU score \citep{SacreBLEU} between the transferred sentence and the source sentence (self-BLEU) or between the transferred sentence and a reference (ref-BLEU) \citep{xu2018unpaired, luo2019a, reif-etal-2022-recipe, suzgun-etal-2022-prompt}. However, BLEU may not accurately score a transferred sentence if it does not have a high n-gram overlap with the reference or the source.

The second category involves extracting the embedding of the transferred sentence, the source, and the reference to calculate their cosine distance. \citet{Fu_Tan_Peng_Zhao_Yan_2018} use pre-trained Glove \citep{pennington-etal-2014-glove} to obtain sentence embeddings, while \citet{roy2023conversation} utilize SBERT \citep{Reimers2019SentenceBERTSE} to extract sentence embeddings. \citet{reid-zhong-2021-lewis} apply the BERTScore approach \citep{Zhang2020BERTScore} to directly evaluate semantic preservation. These approaches can measure the semantic preservation in terms of sentence meaning, which overcomes some of the limitations of using the BLEU score.

In terms of evaluating \textbf{Fluency}, most works employ Perplexity (PPL). \citet{dai-etal-2019-style} and \citet{luo2019a} calculate PPL using 5-gram and pre-trained bi-directional LSTM language models. Recent works prefer to use GPT-2 \citep{radford2019gpt2} to measure PPL \citep{reif-etal-2022-recipe, suzgun-etal-2022-prompt}. 

Along with the prevalence of ChatGPT as an evaluator in other domains \citep{kocmi2023large, wang2023chatgpt}, \citet{lai2023multidimensional} introduce ChatGPT with a designed prompt to evaluate style strength, semantic preservation, and fluency of the style-transferred output simultaneously. Additional related works for TST approaches, subtasks, and datasets can be found in \ref{Additional Related Works} in the Appendix.

\section{Evaluation Methods}

We evaluate two aspects of models in the C-TST setting: style strength and appropriateness, with the latter already accounting for fluency and coherence given contexts.

Regarding style strength, we concentrate on formality and sentiment tasks to establish better links with conventional TST tasks. For the formality task, we employ RoBERTa-base that has been fine-tuned on GYAFC \citep{rao-tetreault-2018-dear} and the online formality corpus \citep{pavlick-tetreault-2016-empirical}. To evaluate the sentiment task, we utilize SiEBERT \citep{HARTMANN202375}.

Concerning appropriateness, the new dimension proposed in this paper, we adopt four approaches to quantify its value: SacreBLEU, Sentence-BERT, ChatGPT and Negative Log Likelihood (NLL). Initially designed for semantic preservation, BLEU and Sentence-BERT can be repurposed in the C-TST setting to evaluate appropriateness, as the reference responses are deemed appropriate. SacreBLEU \citep{SacreBLEU} is a special version of the traditional metric BLEU~\citep{papineni-etal-2002-bleu}, which measures the similarity between model responses and reference samples based on n-gram overlap. Sentence-BERT \citep{Reimers2019SentenceBERTSE} is another conventional measure, which calculates the cosine similarity between the embedding of generated responses and reference responses. However, both metrics suffer from being reference-dependent and unable to handle model responses with significant diversity, which necessitates the two following novel metrics: ChatGPT and NLL. The former instructs ChatGPT to grade responses on a scale of 0-100, while the latter utilizes the Negative Log Likelihood of an open-source LLM to evaluate the model response, with lower Negative Log Likelihood indicating better performance. To make the NLL metric more intuitive, we present it as a positive metric obtained by 100/(Negative Log Likelihood), with higher NLL scores indicating better performance. In our experiments, we adopt Bloom-7b~\citep{workshop2023bloom} as the judge LLM, serving as a proof-of-concept of the effectiveness of NLL metric. It is recommended to use stronger models to obtain more accurate and stable NLL metrics.

Furthermore, we compute the correlation between these four metrics and human evaluation results, providing robust evidence of their validity as reliable metrics. For further details on the evaluation methods, please refer to Appendix~\ref{appendix:lmstyle_benchmark_details}.

\section{Experiments}
\paragraph{Setup} We adopt two widely used dialogue datasets: Daily Dialog \citep{li-etal-2017-dailydialog} and Blended Skill Talk \citep{smith-etal-2020-BST}. All models are instructed to generate responses given single-turn dialogue contexts collected from the first 1,000 samples in both datasets. Additionally, a special prompt is provided for altering the style of responses, allowing for easy switches between multiple styles.

To assess the validity of different metrics, we employ Pearson Correlation and Kendall's Tau Correlation~\citep{Kendall1938} to measure the strength of the connection between these metrics and human evaluation. We only compute sample-level correlations, as they provide more detailed information than system-level correlations:

\begin{center}
$
\begin{array}{r}
 Corr_{sample}=\frac{1}{n} \sum_{i=1}^{n}\left(\rho \left(\left[H_{i,1}, \ldots, H_{i,m}\right]\right.\right., \\
\left.\left.\left[M_{i,1}, \ldots, M_{i,m}\right]\right)\right)
\end{array}
$
\end{center}
Where ${H_{i,j}}$ denotes the human evaluation score of the generated texts of the $j^{th}$ model on $i^{th}$ sample. ${M_{i,j}}$ denotes the metric value of the generated text of the $j^{th}$ model on $i^{th}$ sample. $\rho$ denotes correlation metric including Pearson and Kendall's Tau. For more setup details, please refer to Appendix~\ref{appendix:experimental_details}.

\begin{figure}[h] 
\centering 
\includegraphics[scale=0.2]{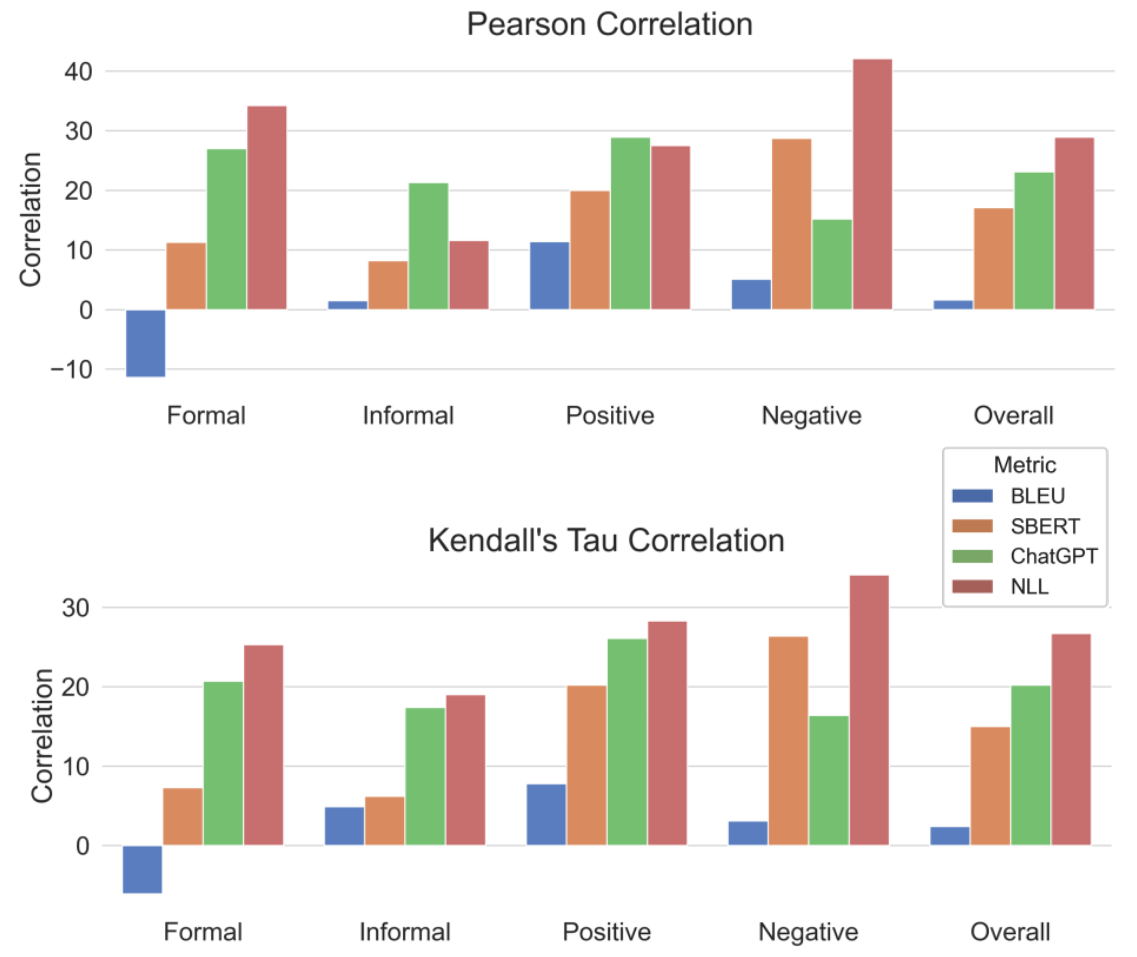} 
\caption{Correlation Result} 
\label{Correlation_plot} 
\end{figure}

\begin{table}[h!]
  \centering
  \resizebox{0.5\textwidth}{!}{\setlength{\tabcolsep}{0.5mm}{
    \begin{tabular}{clccccc}
    \toprule
    \multirow{1}{*}{\; Method\;} & \multicolumn{1}{c}{\multirow{1}{*}{Metric\quad}}
     & \multicolumn{1}{l}{Formal} & \multicolumn{1}{l}{Informal} & \multicolumn{1}{l}{Positive} & \multicolumn{1}{l}{Negative} & \multicolumn{1}{l}{Overall} \\
    \midrule
    \multirow{5}{*}{Pearson} & ACC\% & 36.6 & 22.5 & 16.6 & 32.1 & 27.0 \\
    \cdashline{2-7}
          & BLEU & -11.4 & 1.5 & 11.4 & 5.1 & 1.65 \\
          & SBERT & 11.3 & 8.2 & 20.0 & 28.7 & 17.1 \\
          & ChatGPT & 27.0 & \textbf{21.3} & \textbf{28.9} & 15.2 & 23.1 \\
          & NLL & \textbf{34.2} & 11.6 & 27.5 & \textbf{42.1} & \textbf{28.9} \\
    \midrule
    \multirow{5}{*}{\makecell[c]{Kendall's \\ Tau}} & ACC\% & 32.2 & 16.8 & 17.8 & 19.1 & 21.5 \\
    \cdashline{2-7}
          & BLEU & -6.1 & 4.9 & 7.8 & 3.1 & 2.4 \\
          & SBERT & 7.3 & 6.2 & 20.2 & 26.4 & 15.0 \\
          & ChatGPT & 20.7 & 17.4 & 26.1 & 16.4 & 20.2 \\
          & NLL & \textbf{25.3} & \textbf{19.0} & \textbf{28.3} & \textbf{34.1} & \textbf{26.7}  \\
    \bottomrule
        \end{tabular}%
        }}
\caption{\label{Correlation}
Correlation Results. The result of ACC\% is calculated with human evaluation of style strength. Results of BLEU, SBERT, NLL and ChatGPT are calculated with human evaluation of appropriateness. NLL: a positive metric obtained from 100/(Negative Log Likelihood) using the Bloom-7b1 model.
}
\end{table}%

\begin{table*}[h!]
  \centering
  \resizebox{0.7\textwidth}{!}{\setlength{\tabcolsep}{1mm}{
    \begin{tabular}{llccccccc}
    \toprule
    & \multirow{2}{*}{Model} & \multicolumn{3}{c}{Formality Task} &  & \multicolumn{3}{c}{Sentiment Task} \\
          &  & \multicolumn{1}{l}{ACC\%} & \multicolumn{1}{l}{ChatGPT} & \multicolumn{1}{l}{NLL} & \qquad \qquad & \multicolumn{1}{l}{ACC\%} & \multicolumn{1}{l}{ChatGPT} & \multicolumn{1}{l}{NLL}\\
    \midrule
    & GPT-J-6B & 59.5 & 48.0 & 17.4  &  & 73.2 & 49.0 & 20.0  \\
    & Falcon-7B-Instruct & 54.6 & 55.3 & 33.3  &  & \textbf{87.9} & 50.9 & 30.4 \\
    & RedPajama-7B-Instruct & 54.7 & 69.4 & 37.0  &  & 54.5 & 64.9 & 31.2  \\
    & LLaMA-7B & 53.0 & 86.7 & 32.4 & & 81.2 & 77.4 & 29.2  \\
    & LLaMA-13B & 55.4 & 85.6 & 31.3  & & 80.1 & 75.0 & 29.8  \\
    & Falcon-7B & 52.1 & 80.9 & 38.8  &  & 81.7 & 73.6 & 36.0  \\
    
    & Alpaca-7B & 70.6 & 90.2 & 36.6  &  & 85.6 & 78.9 & 29.0  \\
    & Koala-7B & 77.4 & 80.8 & 35.3  &  & 85.6 & 70.5 & 30.4  \\
    & Koala-13B & 74.4 & 86.5 & 37.8  &  & 84.7 & 79.4 & 34.8  \\
    & Vicuna-7B & \textbf{80.8} & 86.2 & \textbf{41.7}  & & 85.0 & 77.1 & 32.8 \\
    & Vicuna-13B & 77.2 & \textbf{90.3} & 41.4 &  & 82.8 & \textbf{80.0} & \textbf{39.2}  \\
    
    \bottomrule
        \end{tabular}%
        }}
\caption{\label{summary12model}
Comprehensive lists for 11 models' performance on C-TST. ACC\%: Style transfer accuracy for responses. ChatGPT: The appropriateness score evaluated by ChatGPT. NLL: For intuitive purposes, it is presented as a positive metric obtained by 100/(Negative Log Likelihood). 
}
\end{table*}%

\paragraph{Effectiveness of NLL}
As shown in Figure~\ref{Correlation_plot} and Table~\ref{Correlation}, the NLL score achieves the highest correlation with human evaluation of appropriateness across most tasks, with the highest overall correlation for both Pearson and Kendall's Tau methods. This proves the effectiveness of the NLL approach. The ChatGPT and SBERT methods are the second and third most effective approaches in terms of correlation results, while BLEU performs poorly, with correlations on some tasks even being negative. Detailed results can be found in Appendix~\ref{appendix:results_details}.

\paragraph{Comprehensive Results of LLMs}
Table \ref{summary12model} presents a comprehensive list of evaluation results for eleven existing LLMs on the formality task (formal/informal) and sentiment task (positive/negative). It is noteworthy that LLaMA-based models, such as Vicuna, Koala, and Alpaca, all achieve high overall scores. Furthermore, Vicuna exhibits the best overall performance in these C-TST tasks, indicating its strong ability in style adaptation. This also suggests that Vicuna could serve as a valuable starting point for the development of stylized chatbots. For further details and analysis, please refer to Appendix~\ref{appendix:results_details}.

\section{Conclusion}
In this paper, we propose a novel evaluation framework called LMStyle Benchmark, with the aim of providing a standard evaluation paradigm for chat-style text style transfer (C-TST) tasks. In addition to the traditional style strength metrics, LMStyle Benchmark identifies the importance of appropriateness in C-TST evaluation and provides a range of approaches to quantify this new metric. With the support of LMStyle Benchmark, we are able to investigate the stylistic attributes of existing Large Language Models. According to our comprehensive list of results, Vicuna outperforms other models in terms of both formality and sentiment-awareness.

\section*{Limitations}
In this paper, we introduce an automatic evaluation suite for C-TST. We only propose new approaches for appropriateness, but not style strength. However, the evaluation of style strength may become a bottleneck for the overall effectiveness of our evaluation suite.

Additionally, the proposed NLL approach may have some bias if the referee model is similar to the tested model. For this reason, we do not use Vicuna as a referee model, as we need to evaluate several LLaMA-based models. Furthermore, this approach requires a good referee model, which means it has higher hardware requirements than BLEU or SBERT approaches. However, it still costs much less than the ChatGPT approach.

\bibliography{anthology,custom}

\bibliographystyle{acl_natbib}

\appendix

\section{Appendix}
\label{sec:appendix}

\subsection{LMStyle Benchmark in Details}
\label{appendix:lmstyle_benchmark_details}

\subsubsection{ChatGPT Method for Appropriateness Evaluation}
To evaluate appropriateness, we use a specific prompt that instructs ChatGPT to grade generated responses on a scale of 0-100, and then employ regular expressions to extract the score from the output of ChatGPT. The prompt for ChatGPT Evaluation can be found in Figure \ref{GPTPrompt} in the Appendix.

\subsubsection{NLL Method for Appropriateness Evaluation}

Negative Log Likelihood (NLL) is another option for evaluating appropriateness, bearing the core idea of ``predictions reflect real-world knowledge''. The usage of this metric normally involves an addition referee model whose classification ability can be trusted. If a generated output is able to obtain a low-level loss in that model, we may acknowledge to a certain degree that the response is appropriate according to the model.

In particular, to apply the NLL approach in C-TST tasks, we first link the query and the response in the "\#\#\#Speaker: \{\emph{Query}\} \#\#\#Response: \{\emph{Response}\}" format. Next, we use a specific LLM, which we refer to as the referee model, to calculate the NLL of the combined text from the beginning to the end. We then extract the NLL of the \emph{Response} part and divide it by the length of the \emph{Response}, ensuring that longer responses are not evaluated unfairly.

For the referee model, we use Bloom-7b1 \citep{workshop2023bloom}, a relatively small model with strong generation ability. We can load this model with no more than 20G memory, making it more reproducible and easier for us to demonstrate the effectiveness of NLL. However, we strongly recommend employing larger and stronger models as referee models to obtain more stable results.

\subsection{Human Evaluation}
We collect the first 50 queries (source) from Daily Dialog \citep{li-etal-2017-dailydialog} and Blended Skill Talk \citep{smith-etal-2020-BST}, along with their corresponding responses from LLaMA-7B/13B \citep{touvron2023llama} and Vicuna-7B/13B \citep{vicuna2023} models, for human judgement. As a result, we have 100 queries for four tasks, four models, and two dimensions (Style Strength and Appropriateness), resulting in 3,200 assignments for an annotation round.

We recruit four highly proficient workers to annotate them using Amazon Mechanical Turk Sandbox. Each HIT contain four style-transferred responses generated by four models to the same query, making it easier for workers to compare the outputs of different models. We instruct the workers to score the appropriateness and style strength of each response on a continuous scale of 0-100, inspired by \citet{lai-etal-2022-human}. This setting improve the effectiveness of Pearson correlation. Comprehensive instructions with scoring examples are provided to the workers, and after annotating the first 10 HITs, we hold a quality control meeting and provide feedback to clarify any potential misconceptions about the task. The details of the scoring instructions and interface can be found in Figure \ref{Instruction} and \ref{Interface} in the Appendix.

\subsection{Experimental Details}
\label{appendix:experimental_details}
\subsubsection{Data}
As we are evaluating chat-style text style transfer, we collect queries and reference responses from two widely used dialogue datasets, Daily Dialog \citep{li-etal-2017-dailydialog} and Blended Skill Talk \citep{smith-etal-2020-BST}. We choose datasets with references to calculate BLEU and Sentence-BERT (SBERT) for comprehensive analysis. However, it is important to note that our evaluation suite can be used without any references, as the NLL and ChatGPT approaches do not require them.

In this experiment, models generate responses in the single-turn dialogue context. Therefore, we collect the first 1000 dialogue turns separately from the Daily Dialog and Blended Skill Talk (BST) datasets for automatic evaluation. Additionally, we collect all queries and reference responses from the first 1000 dialogues in the two datasets to prevent any manipulation. Similarly, we collect the first 50 query and reference pairs separately from the two datasets for human judgment.

\subsubsection{Models}
In the C-TST setting, employing large language models to generate style-transferred responses with proper prompts is essential, making model selection significant. We select the LLaMA-7B/13B \citep{touvron2023llama} and Vicuna-7B/13B \citep{vicuna2023} models to demonstrate the effectiveness of our automatic evaluation suite, since they are widely recognized as a strong foundation model and a strong fine-tuned model, respectively.

Having proven the effectiveness of our proposed automatic evaluation suite, we can easily evaluate any LLMs. In this paper, we choose to evaluate GPT-J-6b \citep{gpt-j2021}, Alpaca-7B \citep{alpaca2023}, Koala-7B/13B \citep{koala_blogpost_2023}, RedPajama-7B-Instruct \citep{together2023redpajama}, and Falcon-7B/-instruct \citep{falcon2023}. All models, except GPT-J-6B, are kept at a similar 7B/13B level and are released in 2023, making them comparable. We choose not to include BLOOM or BLOOMZ as they are regarded as referee models, which may unfairly score BLOOM-based models. With Vicuna and LLaMA, we have a total of 11 models, which is enough to provide an overview of the performance of current LLMs on C-TST tasks.

\subsubsection{Response Generation}
We follow the prompt-based text style transfer method to generate responses, similar to previous studies such as \citet{reif-etal-2022-recipe}, \citet{suzgun-etal-2022-prompt}, and \citet{roy2023conversation}. Our response generation prompt can be found in Figure \ref{ResponsePrompt} in the Appendix. It allows for easy switching between multiple styles by changing only the style-identification word (Formal/Informal/Positive/Negative), similar to the prompt used in \citet{reif-etal-2022-recipe}. Meanwhile, to construct a generalizable few-shot prompt and avoid over-fitting, we simply collect the style-free few shots from the last 5 dialogues in the evaluation dataset.

\subsubsection{Correlation Methods}
Pearson Correlation takes more information into account, while Kendall's Tau Correlation provides a true pairwise comparison, which can yield more stable results \citep{Vazquez2002}. As we are using human evaluation with a continuous scale setting, we are allowed to calculate both Pearson and Kendall's Tau Correlation \citep{Kendall1938} of various automatic metrics to obtain more robust correlation results.

We only calculate sample-level correlations since they contain more information than system-level correlation. Additionally, we only have human evaluations for four models, making system-level correlation unreliable. The sample-level correlation is calculated by averaging the correlation between the automatic metric and human evaluation score of multiple models' outputs for each individual sample.

\subsection{Results in Details}
\label{appendix:results_details}
\subsubsection{Effectiveness of Automatic Evaluation}
Table \ref{Human} in the Appendix presents the human evaluation results for the generated responses from LLaMA-7b/-13b and Vicuna-7b/-13b on two datasets and four tasks. The results indicate that the responses generated by Vicuna-13b are the most appropriate, followed closely by Vicuna-7b, LLaMA-13b, and LLaMA-7b. This is reasonable since Vicuna models are fine-tuned from LLaMA models using the ShareGPT dialogue dataset, which gives them stronger dialogue abilities. In terms of style strength, Vicuna-7b achieves the highest score, followed closely by Vicuna-13b, while the LLaMA models perform worse. Vicuna-7b outperforms Vicuna-13b in terms of style strength because Vicuna-13b may attach more importance to appropriateness due to the larger training corpus with high appropriateness.

Table \ref{formal4models} and Table \ref{sentiment4models} in the Appendix present the automatic evaluation results of LLaMA-7b/-13b and Vicuna-7b/-13b on different datasets and tasks. Each task and dataset contains 1,000 queries and 4,000 responses from four models. Consequently, each metric evaluates 32,000 responses, enhancing the robustness of the results. The results show that Sentence-BERT (SBERT), NLL score, and ChatGPT score generally give the best scores to Vicuna-13b, while BLEU cannot correctly recognize the performance of Vicuna-13b. Among SBERT, NLL score, and ChatGPT score, NLL score is most stable because it never scores LLaMA models as the best, unlike SBERT and ChatGPT.

\subsubsection{Correlation Results}

To further analyze the automatic evaluation metrics, we conduct a comprehensive correlation analysis using Pearson and Kendall's Tau methods on different tasks. As shown in Table \ref{Correlation} and Figure \ref{Correlation_plot}, the NLL score achieves the highest correlation with human evaluation of appropriateness on most tasks and the highest overall correlation for both Pearson and Kendall's Tau methods. This proves the effectiveness of the NLL method. The ChatGPT and SBERT approaches are the second and third best approaches regarding correlation results, while BLEU performs poorly, with correlations on some tasks even being negative. These correlation results are consistent with the previous results derived from Table \ref{formal4models} and Table \ref{sentiment4models}. However, the correlation results cannot completely reject the ChatGPT approach. ChatGPT tends to score responses with slight differences equally, which may cause a low correlation value, especially under the Kendall's Tau scheme. Hence, we select both NLL score and ChatGPT score for our new evaluation suite to measure response appropriateness.

The overall correlation between ACC\% and human evaluation of style strength is about 25\%, which is fair but not very good. This is likely because we do not use different fine-grained few-shot prompts for each task. The style understood by the style classifier may be somewhat different from the style understood by the LLMs. We do not use fine-grained few-shot prompts for different tasks because it reduces the generalization ability of prompt-based text style transfer approaches \citep{reif-etal-2022-recipe}. Meanwhile, few other methods are available to the evaluation of style strength \citep{lai-etal-2022-human}. However, the ACC\% metric is sufficient for our purposes, as the key focus of this paper is the evaluation of appropriateness.

\subsubsection{Results of Various LLMs}
Table \ref{summary12model} presents the overall performance of 11 models on the formality task (formal/informal) and sentiment task (positive/negative). We evaluate them on the combined dataset of Daily Dialog and Blended Skill Talk. The detailed results of these models' performance on different subtasks and datasets can be found in Tables \ref{formal12model} and \ref{sent12model} in the Appendix.

We can see that models fine-tuned on top of LLaMA, such as Vicuna, Koala, and Alpaca, achieve high overall scores, demonstrating the high potential of the LLaMA foundation. Vicuna models show the best overall performance in these C-TST tasks, indicating the high quality of the fine-tuning process in Vicuna. It is also reasonable that GPT-J-6B performs the worst since it is the earliest foundation model.

Table \ref{summary12model} also reveals some interesting results. For example, the responses generated by Falcon-7B are much more appropriate than those generated by Falcon-7B-Instruct. One explanation is that the instruct model places too much emphasis on style at the expense of appropriateness, as evidenced by the results of the sentiment task. 

\subsection{Additional Related Works} \label{Additional Related Works}
\textbf{TST Approaches.} \ Previous works mainly focus on introducing new methods to outperform in the TST task. Some researchers, such as \citet{xu-etal-2012-paraphrasing} and \citet{rao-tetreault-2018-dear}, apply supervised training methods on parallel datasets. For non-parallel datasets, one of the widely used approaches involves disentangling the representation of style-independent content and style and fusing the content representation with desired styles to generate style-transferred texts \citep{shen2017, prabhumoye-etal-2018-style, Fu_Tan_Peng_Zhao_Yan_2018, luo2019a}. Another line of research is devoted to identifying and replacing style-related phrases to solve TST tasks \citep{li-etal-2018-delete, wu2019, malmi-etal-2020-unsupervised, reid-zhong-2021-lewis}.

Along with the development of large language models (LLMs), prompt-based approaches are introduced in recent works \citep{reif-etal-2022-recipe, suzgun-etal-2022-prompt, luo2023promptbased, roy2023conversation}. These approaches remove the need for a training process or labeled data, allowing for their extension to arbitrary style transfer tasks. It is a promising field, but no work has been done on the comprehensive evaluation of different advanced LLMs, such as LLaMA \citep{touvron2023llama} and Vicuna \citep{vicuna2023}, using prompt-based approaches on various TST tasks. Our research attempts to fill this gap.

\textbf{Datasets and Subtasks.} While prompt-based methods do not require parallel data, having sufficient source-reference pairs for evaluation is desirable. The YELP review dataset \citep{zhang2015} is the most widely used data in TST for sentiment transfer tasks \citep{wu2019, luo2019a, Luo19b, dai-etal-2019-style, malmi-etal-2020-unsupervised, goyal-etal-2021-multi}, and the AMAZON product review dataset \citep{He2016} is another popular dataset that has been further annotated by \citet{li-etal-2018-delete} for sentiment transfer tasks \citep{li-etal-2018-delete, Fu_Tan_Peng_Zhao_Yan_2018,luo2023promptbased}.

Formality transfer is another important TST subtask, with the GYAFC dataset \citep{rao-tetreault-2018-dear} being the most widely used dataset for this task \citep{reif-etal-2022-recipe, luo2023promptbased}. Additionally, \citet{Madaan2020} and \citet{reid-zhong-2021-lewis} use the POLITE dataset to control the politeness of generated texts, which is similar to the formality transfer task.

In this paper, we also focus on formality and sentiment TST tasks. However, we do not use datasets such as YELP and GYAFC, as they are only suitable for T-TST tasks. For the C-TST task, which requires dialogue data, we collect data from the Daily Dialog \citep{li-etal-2017-dailydialog} and Blended Skill Talk \citep{smith-etal-2020-BST} datasets, inspired by \citet{han-etal-2022-meet}.

\begin{figure*}[h] 
\centering 
\includegraphics[scale=0.4]{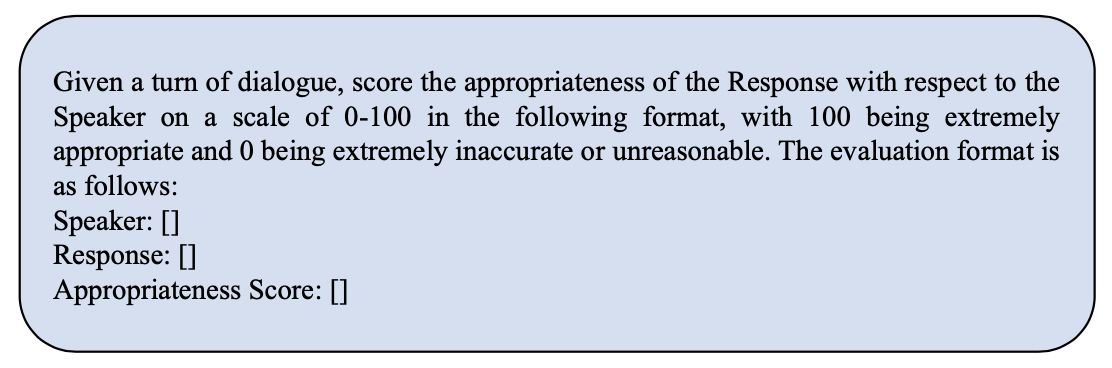} 
\caption{Prompt for ChatGPT Evaluation} 
\label{GPTPrompt} 
\end{figure*}

\begin{figure*}[h] 
\centering 
\includegraphics[scale=0.5]{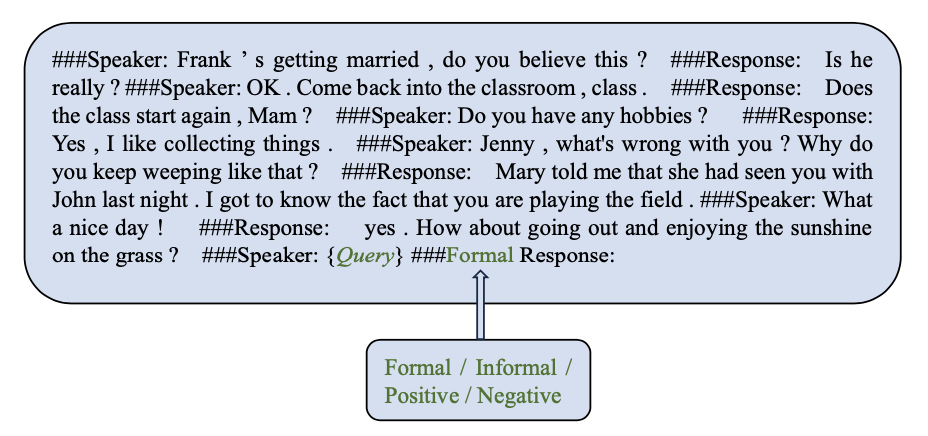} 
\caption{Prompt for Style-Transferred Response Generation} 
\label{ResponsePrompt} 
\end{figure*}

\begin{table*}[htbp]
  \centering
  \resizebox{\textwidth}{!}{\setlength{\tabcolsep}{1.2mm}{
    \begin{tabular}{lccccccccccc}
    \toprule
    \multicolumn{1}{c}{\multirow{2}[2]{*}{Model}} & \multicolumn{5}{c}{Appropriateness}   &       & \multicolumn{5}{c}{Style Strength} \\
          & \multicolumn{1}{l}{Formal} & \multicolumn{1}{l}{Informal} & \multicolumn{1}{l}{Positive} & \multicolumn{1}{l}{Negative} & \multicolumn{1}{l}{Overall} &   \qquad \qquad    & \multicolumn{1}{l}{Formal} & \multicolumn{1}{l}{Informal} & \multicolumn{1}{l}{Positive} & \multicolumn{1}{l}{Negative} & \multicolumn{1}{l}{Overall} \\
    \midrule
    LLaMA-7B & 45.2  & 55.4  & 55.3  & 51.1  & 51.8 &       & 52.9  & 42.6  & 63.4  & 66.0    & 56.2 \\
    LLaMA-13B & 52.5  & 53.5  & 59.2  & 54.3  & 54.9 &       & 56.7  & 52.4  & 64.9  & \textbf{68.3}  & 60.6 \\
    Vicuna-7B & 65.0    & 62.9  & 68.9  & 59.5  & 64.1 &       & \textbf{73.2}  & \textbf{71.5}  & 66.4  & 66.5  & \textbf{69.4} \\
    Vicuna-13B & \textbf{79.6}  & \textbf{73.2}  & \textbf{77.5}  & \textbf{71.1}  & \textbf{75.4} &       & 68.8  & 67.4  & \textbf{70.3}  & 62.5  & 67.3 \\
    \bottomrule
    \end{tabular}%
    }}
\caption{\label{Human}
Human Evaluation Results. The Appropriateness/Style score of each model is calculated by averaging the annotated Appropriateness/Style score of model-generated texts.
}
\end{table*}%

\begin{table*}[htbp]
  \centering
  \resizebox{\textwidth}{!}{\setlength{\tabcolsep}{1.2mm}{
    \begin{tabular}{clccccccccccc}
    \toprule
    \multirow{2}{*}{Dataset} & \multicolumn{1}{c}{\multirow{2}{*}{Model}} & \multicolumn{5}{c}{Formal} &  & \multicolumn{5}{c}{Informal} \\
          &       & \multicolumn{1}{l}{ACC\%} & \multicolumn{1}{l}{BLEU} & \multicolumn{1}{l}{SBERT} & \multicolumn{1}{l}{NLL} & \multicolumn{1}{l}{ChatGPT} & \qquad \qquad & \multicolumn{1}{l}{ACC\%} & \multicolumn{1}{l}{BLEU} & \multicolumn{1}{l}{SBERT} & \multicolumn{1}{l}{NLL} & \multicolumn{1}{l}{ChatGPT} \\
    \midrule
    \multirow{4}{*}{Daily Dialog} & LLaMA-7B & 12.1 & 5.1 & 28.9 & 34.8 & 90.6 &   & 94.2 & 6.3 & 29.6 & 33.3 & 87.6 \\
          & LLaMA-13B & 12.4 & \textbf{6.8} & 30.6 & 35.0 & 91.6 &   & \textbf{95.1} & \textbf{6.6} & 29.5 & 28.9 & \textbf{88.5}  \\
          & Vicuna-7B & 67.1 & 5.3 & 33.1 & \textbf{47.0} & 93.7 &   & 94.1 & 4.7 & 30.2 & \textbf{35.8} & 82.7 \\
          & Vicuna-13B & \textbf{68.4} & 5.1 & \textbf{33.5} & 46.8 & \textbf{94.6} &   & 87.9 & 5.4 & \textbf{30.6} & 34.8 & 87.8 \\
    \midrule
    \multirow{4}{*}{Blended Skill Talk} & LLaMA-7B & 23.2 & 3.2 & 26.3 & 31.5 & 86.5  &   & 82.3 & 3.1 & 26.5 & 29.8 & 82.1 \\
          & LLaMA-13B & 31.0 & \textbf{3.4} & 26.7 & 33.8 & 83.6 &   & \textbf{83.3} & 3.0 & 24.9 & 27.5 & 78.8 \\
          & Vicuna-7B & \textbf{95.0} & 3.2 & 30.6 & \textbf{48.0} & 87.6 &   & 66.9 & 3.3 & 28.6 & 35.8 & 80.7 \\
          & Vicuna-13B & 94.6 & 3.1 & \textbf{34.0} & 47.8 & \textbf{93.0} &   & 57.8 & \textbf{3.4} & \textbf{30.0} & \textbf{36.3} & \textbf{85.9} \\
    \bottomrule
        \end{tabular}%
        }}
\caption{\label{formal4models}
Results on formality tasks. ACC\%: Style transfer accuracy for Formal/Informal responses. NLL: a positive metric obtained from 100/(Negative Log Likelihood) using the Bloom-7b1 model. ChatGPT: The appropriateness score evaluated by ChatGPT.
}
\end{table*}%

\begin{table*}[htbp]
  \centering
  \resizebox{\textwidth}{!}{\setlength{\tabcolsep}{1.2mm}{
    \begin{tabular}{clccccccccccc}
    \toprule
    \multirow{2}{*}{Dataset} & \multicolumn{1}{c}{\multirow{2}{*}{Model}} & \multicolumn{5}{c}{Positive} &  & \multicolumn{5}{c}{Negative} \\
          &       & \multicolumn{1}{l}{ACC\%} & \multicolumn{1}{l}{BLEU} & \multicolumn{1}{l}{SBERT} & \multicolumn{1}{l}{NLL} & \multicolumn{1}{l}{ChatGPT} & \qquad \qquad & \multicolumn{1}{l}{ACC\%} & \multicolumn{1}{l}{BLEU} & \multicolumn{1}{l}{SBERT} & \multicolumn{1}{l}{NLL} & \multicolumn{1}{l}{ChatGPT} \\
    \midrule
    \multirow{4}{*}{Daily Dialog} & LLaMA-7B & 76.1 & 6.7 & 29.7 & 32.3 & 92.0 &   & 85.7 & \textbf{5.2} & 24.0 & 31.9 & \textbf{72.7} \\
          & LLaMA-13B & 72.7 & \textbf{7.1} & 30.4 & 30.9 & 93.0 &   & \textbf{88.3} & 5.0 & 23.8 & 32.3 & 69.8 \\
          & Vicuna-7B & \textbf{81.6} & 6.3 & 31.9 & 36.5 & 94.0 &   & 87.9 & 4.7 & 23.6 & 33.3 & 71.5 \\
          & Vicuna-13B & 79.2 & 5.9 & \textbf{33.8} & \textbf{38.0} & \textbf{95.3} &   & 86.8 & 4.2 & \textbf{27.2} & \textbf{39.5} & 72.0 \\
    \midrule
    \multirow{4}{*}{Blended Skill Talk} & LLaMA-7B & 72.0 & 3.2 & 24.4 & 26.4 & 87.1 &   & 90.8 & 2.9 & 20.3 & 26.0 & 57.7 \\
          & LLaMA-13B & 68.1 & 3.2 & 25.3 & 27.9 & 85.2 &   & 91.2 & 3.0 & 20.7 & 28.1 & 52.0 \\
          & Vicuna-7B & 78.1 & 3.2 & 27.8 & 31.3 & 87.9 &   & \textbf{92.5} & 2.9 & 20.3 & 30.3 & 54.9 \\
          & Vicuna-13B & \textbf{81.8} & \textbf{3.4} & \textbf{32.8} & \textbf{40.0} & \textbf{91.1} &   & 83.4 & \textbf{3.0} & \textbf{25.6} & \textbf{39.5} & \textbf{61.7} \\
    \bottomrule
        \end{tabular}%
        }}
\caption{\label{sentiment4models}
Results on sentiment tasks. ACC\%: Style transfer accuracy for responses. ChatGPT: The appropriateness score evaluated by ChatGPT. NLL: For intuitive purposes, it is presented as a positive metric obtained by 100/(Negative Log Likelihood).
}
\end{table*}%

\begin{table*}[htbp]
  \centering
  \resizebox{0.8\textwidth}{!}{\setlength{\tabcolsep}{1mm}{
    \begin{tabular}{clccccccc}
    \toprule
    \multirow{2}{*}{Dataset} & \multicolumn{1}{c}{\multirow{2}{*}{Model}} & \multicolumn{3}{c}{Formal} &  & \multicolumn{3}{c}{Informal} \\
          &       & \multicolumn{1}{l}{ACC\%} & \multicolumn{1}{l}{ChatGPT} & \multicolumn{1}{l}{NLL} & \qquad \qquad & \multicolumn{1}{l}{ACC\%} & \multicolumn{1}{l}{ChatGPT} & \multicolumn{1}{l}{NLL} \\
    \midrule
    \multirow{11}{*}{Daily Dialog}
    & GPT-J-6B & 60.7 & 69.3 & 24.3  &  & 56.5 & 61.3 & 19.8  \\
    & Falcon-7B-Instruct & \textbf{92.2} & 69.6 & 39.5  &  & 13.6 & 57.4 & 31.0 \\
    & RedPajama-7B-Instruct & 69.5 & 74.6 & 38.5 &  & 36.0 & 77.6 & 39.3 \\
    & LLaMA-7B & 12.1 & 90.6 & 34.8  & & 94.2 & 87.6 & 33.3 \\
    & LLaMA-13B & 12.4 & 91.6 & 35.0 & & 95.1 & 88.5 & 28.9  \\
    & Falcon-7B & 88.7 & 87.9 & 42.3  &  & 17.5 & 87.4 & \textbf{39.8}  \\
    & Alpaca-7B & 44.6 & \textbf{94.9} & 40.0  &  & 94.4 & 87.7 & 35.3  \\
    & Koala-7B & 55.0 & 87.8 & 42.5  &  & \textbf{95.9} & 82.9 & 34.8 \\
    & Koala-13B & 56.9 & 94.2 & 42.5 &  & 95.3 & \textbf{88.8} & 33.0 \\
    & Vicuna-7B & 67.1 & 93.7 & \textbf{47.0}  & & 94.1 & 82.7 & 35.8  \\
    & Vicuna-13B & 68.4 & 94.6 & 46.8  &  & 87.9 & 87.8 & 34.8   \\
    
    \midrule
    \multirow{11}{*}{Blended Skill Talk} 
    & GPT-J-6B & 35.2 & 39.5 & 14.1 &  & \textbf{85.5} & 22.0 & 11.4 \\
    & Falcon-7B-Instruct & \textbf{96.6} & 54.7 & 36.5 &  & 16.1 & 39.5 & 26.1 \\
    & RedPajama-7B-Instruct & 63.0 & 60.6 & 36.5 &  & 50.4 & 65.0 & 33.8 \\
    & LLaMA-7B & 23.2 & 86.5 & 31.5  & & 82.3 & 82.1 & 29.8 \\
    & LLaMA-13B & 31.0 & 83.6 & 33.8  & & 83.3 & 78.8 & 27.5 \\
    & Falcon-7B & 85.1 & 73.1 & 37.3 &  & 17.0 & 75.3 & 35.8 \\
    & Alpaca-7B & 66.9 & 91.5 & 39.3  &  & 76.4 & \textbf{86.5} & 31.6 \\
    & Koala-7B & 77.0 & 73.0 & 36.8 &  & 81.9 & 79.3 & 27.0 \\
    & Koala-13B & 86.2 & 84.1 & 42.5 &  & 59.3 & 78.8 & 33.0 \\
    & Vicuna-7B & 95.0 & 87.6 & \textbf{48.0}  & & 66.9 & 80.7 & 35.8 \\
    & Vicuna-13B & 94.6 & \textbf{93.0} & 47.8  &  & 57.8 & 85.9 & \textbf{36.3} \\
    
    \bottomrule
        \end{tabular}%
        }}
\caption{\label{formal12model}
Comprehensive lists for 11 models on formality tasks. ACC\%: Style transfer accuracy for responses. ChatGPT: The appropriateness score evaluated by ChatGPT. NLL: For intuitive purposes, it is presented as a positive metric obtained by 100/(Negative Log Likelihood). 
}
\end{table*}%

\begin{table*}[htbp]
  \centering
  \resizebox{0.8\textwidth}{!}{\setlength{\tabcolsep}{1mm}{
    \begin{tabular}{clccccccccc}
    \toprule
    \multirow{2}{*}{Dataset} & \multicolumn{1}{c}{\multirow{2}{*}{Model}} & \multicolumn{3}{c}{Positive} &  & \multicolumn{3}{c}{Negative} \\
          &       & \multicolumn{1}{l}{ACC\%} & \multicolumn{1}{l}{ChatGPT} & \multicolumn{1}{l}{NLL}  & \qquad \qquad & \multicolumn{1}{l}{ACC\%} & \multicolumn{1}{l}{ChatGPT} & \multicolumn{1}{l}{NLL} \\
    \midrule
    \multirow{11}{*}{Daily Dialog}
    & GPT-J-6B & 70.6 & 72.1 & 25.3  &  & 91.9 & 48.8 & 23.9 \\
    & Falcon-7B-Instruct & \textbf{86.2} & 69.1 & 33.2 &  & 90.2 & 50.7 & 35.5 \\
    & RedPajama-7B-Instruct & 34.7 & 70.5 & 32.5 &  & 77.9 & 69.0 & 36.8 \\
    & LLaMA-7B & 76.1 & 92.0 & 32.3  & & 85.7 & 72.7 & 31.9 \\
    & LLaMA-13B & 72.7 & 93.0 & 30.9  & & 88.3 & 69.8 & 32.3  \\
    & Falcon-7B & 72.9 & 91.8 & \textbf{38.8} &  & 86.3 & 69.8 & 39.0 \\
    & Alpaca-7B & 84.3 & 93.5 & 33.5  &  & 88.9 & 72.8 & 30.1 \\
    & Koala-7B & 76.0 & 89.6 & 36.0  &  & \textbf{93.8} & 59.3 & 32.3 \\
    & Koala-13B & 79.7 & 94.8 & 36.5 &  & 87.6 & \textbf{72.9} & 34.0 \\
    & Vicuna-7B & 81.6 & 94.0 & 36.5  & & 87.9 & 71.5 & 33.3 \\
    & Vicuna-13B & 79.2 & \textbf{95.3} & 38.0 &  & 86.8 & 72.0 & \textbf{39.5}  \\
    
    \midrule
    \multirow{11}{*}{Blended Skill Talk} 
    & GPT-J-6B & 33.9 & 40.2 & 14.3 &  & 96.2 & 35.0 & 16.5 \\
    & Falcon-7B-Instruct & 78.1 & 54.0 & 23.8 &  & \textbf{97.2} & 29.7 & 29.1 \\
    & RedPajama-7B-Instruct & 20.5 & 55.4 & 26.1 &  & 84.8 & \textbf{64.8} & 29.3 \\
    & LLaMA-7B & 72.0 & 87.1 & 26.4 & & 90.8 & 57.7 & 26.0 \\
    & LLaMA-13B & 68.1 & 85.2 & 27.9 & & 91.2 & 52.0 & 28.1 \\
    & Falcon-7B & 72.2 & 81.8 & 33.5 &  & 95.4 & 50.8 & 32.5 \\
    & Alpaca-7B & 81.0 & 89.0 & 27.5 &  & 88.2 & 60.4 & 25.0 \\
    & Koala-7B & 76.4 & 81.9 & 26.0 &  & 96.0 & 51.2 & 27.5 \\
    & Koala-13B & 80.3 & 87.6 & 36.0 &  & 91.3 & 62.5 & 32.8 \\
    & Vicuna-7B & 78.1 & 87.9 & 31.3 & & 92.5 & 54.9 & 30.3 \\
    & Vicuna-13B & \textbf{81.8} & \textbf{91.1} & \textbf{40.0} &  & 83.4 & 61.7 & \textbf{39.5} \\
    
    \bottomrule
        \end{tabular}%
        }}
\caption{\label{sent12model}
Comprehensive lists for 11 models on sentiment tasks. ACC\%: Style transfer accuracy for responses. ChatGPT: The appropriateness score evaluated by ChatGPT. NLL: For intuitive purposes, it is presented as a positive metric obtained by 100/(Negative Log Likelihood).
}
\end{table*}%

\begin{figure*}[h] 
\centering 
\includegraphics[scale=0.6]{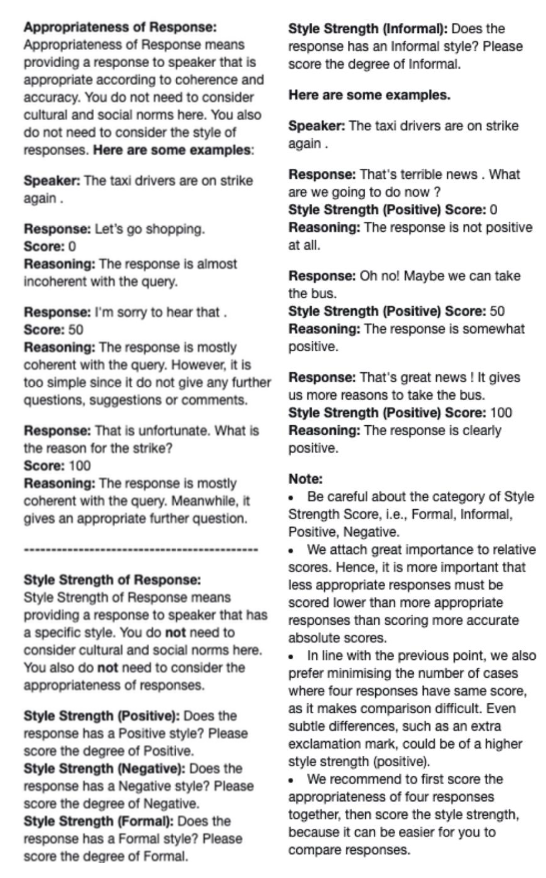} 
\caption{Scoring Instruction for Human Evaluation} 
\label{Instruction} 
\end{figure*}

\begin{figure*}[h] 
\centering 
\includegraphics[scale=0.4]{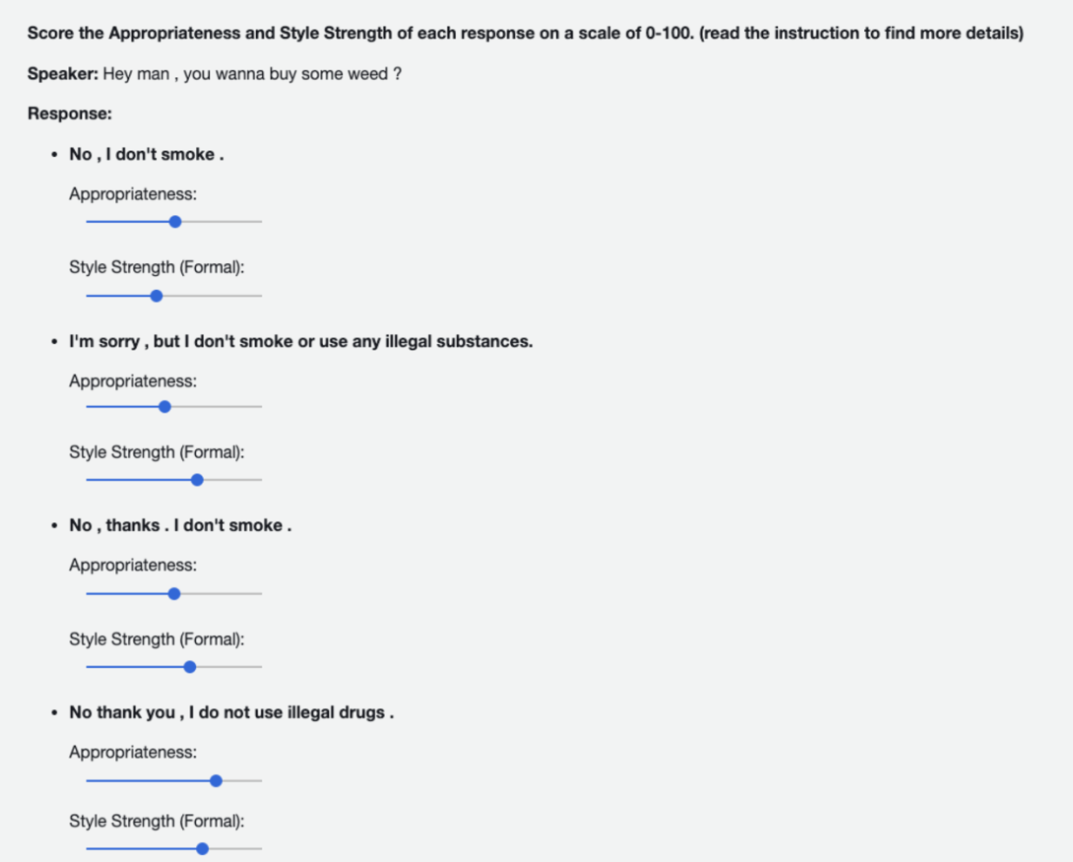} 
\caption{Scoring Interface for Human Evaluation} 
\label{Interface} 
\end{figure*}

\end{document}